\documentclass{article}
\usepackage[numbers]{natbib}

\usepackage[final,main]{neurips_2026}

\usepackage[T1]{fontenc}    
\usepackage{url}            
\usepackage{booktabs}       
\usepackage{amsfonts}       
\usepackage{nicefrac}       
\usepackage{microtype}      
\usepackage{xcolor}         
\usepackage{hyperref}       
\usepackage{amsmath}
\usepackage{graphicx}
\usepackage{algorithm}
\usepackage{algpseudocode}
\usepackage{subcaption}
\usepackage{listings}
\usepackage{xcolor}
\usepackage{float}
\usepackage{caption}
\usepackage{xcolor}
\usepackage[utf8]{inputenc}

\usepackage[colorinlistoftodos]{todonotes}

\lstdefinestyle{pythonstyle}{
    language=Python,
    basicstyle=\ttfamily\small,
    keywordstyle=\bfseries,
    commentstyle=\itshape,
    stringstyle=\ttfamily,
    showstringspaces=false,
    breaklines=true,
    frame=single,
    columns=fullflexible,
    keepspaces=true,
    tabsize=4
}

\title{
New Bounds for Zarankiewicz Numbers via Reinforced LLM Evolutionary Search\\
{\large \textcolor{red}{Working Paper}}
}

\author{%
Jay~Bhan\textsuperscript{*}\\
Massachusetts Institute of Technology\\
Cambridge, MA 02139 \\
\texttt{jaybhan@mit.edu} \\
\And
Nicole~Nobili\equalcontrib\\
ETH Zürich\\
Zürich, Switzerland 8092\\
\texttt{nnobili@ethz.ch} \\
\And
Srinivasan~Raghuraman\\
Massachusetts Institute of Technology\\
Cambridge, MA 02139\\
\texttt{srirag@mit.edu} \\
\And
Patrick~Langer\\
Agentic Systems Lab, ETH Zürich\\
Zürich, Switzerland 8092\\
Stanford University\\
Stanford, CA 94305\\
\texttt{planger@stanford.edu} \\
}

\begin{document}
\setcitestyle{numbers}

\newcommand{\equalcontrib}{\thanks{These authors contributed equally to this work as first authors, and their order was determined via coin flip.}}

\maketitle
\begin{abstract}
The Zarankiewicz number $\textbf{Z}(m, n, s, t)$ is the maximum number of edges in a bipartite graph $G_{m, n}$ such that there is no complete $K_{s, t}$ bipartite subgraph. We determine for the first time the exact values of three Zarankiewicz numbers: $\textbf{Z}(11, 21, 3, 3)=116$, $\textbf{Z}(11, 22, 3, 3)=121$, and $\textbf{Z}(12, 22, 3, 3)=132$. We further establish lower bounds for 41 more Zarankiewicz numbers, including several that are within one edge of the best known upper bound, and we match the established value in four more closed cases. Our results are obtained using OpenEvolve, an open-source evolutionary algorithm based on Large Language Models (LLMs) that iteratively improves algorithms for generating mathematical constructions by optimizing a reward signal which we tailored for this specific problem. These findings provide new extremal graph constructions and demonstrate the potential of LLM-guided evolutionary search to contribute to mathematical research. In addition to presenting the resulting constructions, we report the generation algorithms produced, describe the relevant implementation details, and provide our computational costs. Our costs are remarkably low, at less than \$30 for each Zarankiewicz parameter combination, showing that LLM-guided evolutionary search can be an inexpensive, reproducible, and accessible tool for discovering new combinatorial constructions.
\end{abstract}

\section{Introduction}
\label{sec:intro}
The Zarankiewicz problem concerns the computation of the Zarankiewicz numbers $\textbf{Z}(m, n, s, t)$ defined to be the maximum number of edges in a bipartite graph with parts of size $m$ and $n$ that avoids a copy of $K_{s,t}$, the complete bipartite graph with parts of size $s$ and $t$.
This problem lies at the core of extremal graph theory. The Zarankiewicz numbers capture, in a precise quantitative form, the principle that sufficient density forces the appearance of forbidden structures. One can also view the Zarankiewicz problem as the bipartite analogue of classical Tur\'an-type problems \cite{turan1941extremal}. The foundational K\H{o}v\'ari–S\'os–Tur\'an theorem from 1954 \cite{kovari1954zarankiewicz} establishes the upper bound $\textbf{Z}(m, n, s, t) = O(n^{2 - 1/s})$, and despite extensive work, determining the correct asymptotics for fixed $s$ and $t$ remains open.

In keeping with extremal graph theory, Zarankiewicz upper bounds translate directly into upper bounds on bipartite Ramsey numbers \cite{goddard2000bipartite, collins2016zarankiewicz}, and constructions of dense $K_{s,t}$-free graphs yield corresponding lower bounds. Beyond extremal graph theory, Zarankiewicz-type bounds additionally play a structural role in the analysis of Boolean matrices. Identifying a bipartite graph with its adjacency matrix $A \in \{0, 1\}^{m \times n}$, forbidding $K_{s,t}$ is equivalent to forbidding an all-ones $s\times t$ submatrix. This formulation has applications to several areas of theoretical computer science. For instance, in communication complexity, this perspective is central to the so-called rectangle covering method, i.e., extremal upper bounds on $\textbf{Z}(m, n, s, t)$ translate directly into communication lower bounds \cite{Hayes2011Separating}. In circuit complexity, the same combinatorial constraint governs depth-$2$ monotone representations of Boolean matrices \cite{jukna2013complexity}. There are many more applications to areas such as property testing, additive combinatorics, incidence geometry, pseudorandomness, and cryptography  \cite{networkot, Zhao2023GTAC, ForeyFresanKowalskiWigderson2025, Smorodinsky2025SurveyZarankiewicz} 

 Our work focuses on creating constructions confirming lower bounds for Zarankiewicz numbers. To do so, we implement an algorithm based on an open-source implementation of Google DeepMind's AlphaEvolve \cite{novikov2025alphaevolve}, OpenEvolve \cite{openevolve}. AlphaEvolve's structure revolves around a system of LLMs that develop algorithms for solving a mathematical problem, which are continuously improved by the feedback received from previous rounds. Recently, DeepMind used AlphaEvolve to tighten lower bounds on seven previously unknown Ramsey numbers, marking a great milestone in AI's ability to solve combinatorial problems \cite{nagda2026ramsey}. Inspired by this recent work, we investigate whether the Zarankiewicz numbers share a similar combinatorial structure that this architecture succeeds with. We show that it does.

\section{Method}
\subsection{Formulating the problem}
\label{subsec:problem_formulation}

As stated in Section~\ref{sec:intro}, the Zarankiewicz number $\textbf{Z}(m, n, s, t)$ is defined as the maximum number of edges in a bipartite graph with parts of size $m$ and $n$ that contains no complete bipartite subgraph $K_{s,t}$ of size $s$ and $t$. Thus, we can establish a $c$ as a \textit{lower bound}, $\textbf{Z}(m, n, s, t) \geq c$, for the Zarankiewicz number $\textbf{Z}(m, n, s, t)$ if we can find a \textit{construction} $G_{m, n}$ with $c$ edges and no complete $K_{s, t}$ bipartite subgraph.

The first step in applying evolutionary search guided by LLMs to the Zarankiewicz problem is to choose an effective representation of the construction defined above. Since our method uses LLMs to generate candidate solutions, we are looking for a formulation of the construction that LLMs perform well with.

Although the Zarankiewicz problem is often stated in terms of graphs, recent benchmarks show that direct graph reasoning remains challenging for current LLMs. Performance varies across graph types and graph concepts, and even strong models struggle on larger and more complex graph computation problems~\cite{tang2025grapharena,wu2025grapheval36k}. We therefore avoid representing candidates directly as graphs. Instead, we use a formulation based on matrices. As introduced in Section~\ref{sec:intro}, the Zarankiewicz number $\mathbf{Z}(m,n,s,t)$ can in fact be equivalently defined as the largest number of ones you can fit into a binary $m{\times}n$ matrix such that the intersection of any $s$ rows and $t$ columns does not contain exclusively ones.

Following recent work applying AI to extremal combinatorics, such as \cite{nagda2026ramsey,nagda2025reinforced,charton2024patternboost}, we do not use the LLM to generate matrices directly. Instead, we use it to evolve algorithms for constructing binary $m \times n$ matrices. Starting from previously discovered candidate algorithms, the search iteratively proposes mutations to previously discovered candidate algorithms with the goal of producing matrices with more ones while preserving all constraints on forbidden submatrices.

\subsection{AlphaEvolve}
AlphaEvolve \cite{novikov2025alphaevolve} is an evolutionary algorithm introduced by Google DeepMind for making algorithmic discoveries. AlphaEvolve maintains a population of candidate programs, uses LLMs to propose mutations or refinements, evaluates the resulting programs with a scoring function, and feeds the best candidates back into later generations for further evolution. Unlike approaches that adapt model weights to a specific problem, such as TTT-discover \cite{yuksekgonul2026learning}, the discovery process happens entirely at inference time, leveraging the existing code generation and reasoning capabilities of pretrained LLMs without training on the target problem. We base our implementation on OpenEvolve, an open source implementation inspired by AlphaEvolve.

We choose to adopt an AlphaEvolve style approach for two main reasons: firstly, we are motivated by the recent successes of applying evolutionary methods driven by LLMs to extremal combinatorics \cite{nagda2026ramsey,nagda2025reinforced}. Secondly, the recent successes from AlphaEvolve and related work \cite{novikov2025alphaevolve, yuksekgonul2026learning, openevolve} testify how evolutionary search driven by LLMs is particularly effective when the goal is to discover complex constructions. These successes motivate our use of the same paradigm for the Zarankiewicz problem, where progress likewise depends on discovering a construction, more specifically matrices with many ones that satisfy exact constraints on forbidden submatrices.

\subsection{Choice of Zarankiewicz parameters}
Within the family of Zarankiewicz numbers $\textbf{Z}(m,n,s,t)$, we focus on the case $s=t=3$ for the following reasons. First, the diagonal case $s=t$ is especially natural because it imposes the same forbidden complete bipartite subgraph condition on both parts of the bipartition. By eliminating asymmetry between the two sides, this setting provides a canonical benchmark family. Second, within the diagonal family, the case $s=t=2$ is the first nontrivial instance and has been studied extensively \cite{tan2022zarankiewiczsat, davies2026zarankiewicz}. The diagonal case $s=t=3$, is therefore a natural next target. Within this setting, we focus on the previously unresolved range $9 \leq m \leq 16$ and $17 \leq n \leq 23$, which lies immediately adjacent to the region of established values. In $9 \leq m \leq 16$ and $17 \leq n \leq 23$, we focus on 44 combinations of $m$ and $n$ for which the $s=t=3$ Zarankiewicz number has not been previously established (see \autoref{fig:zar_nums}). Within this range, we also run experiments on seven other combinations of $m \times n$ where the Zarankiewicz number has previously been established (see \autoref{fig:zar_nums}). We thus run experiments for 51 $m \times n$ combinations total.

\subsection{Scoring algorithm}
Motivated by \citeauthor{nagda2026ramsey}'s application of AlphaEvolve to Ramsey numbers \cite{nagda2026ramsey}, we adopt the search algorithm detailed in \autoref{fig:Alg}.

For every tuple of $(m,n)$, we initialize the algorithm with a program $\mathcal{P}$ that outputs a sparsely populated binary matrix, and with a Zarankiewicz lower bound $n_{\text{SoTA}}$ of 0.

Within each iteration of the program search, an algorithm is selected from the current population of programs based on its score and then mutated. During the mutation process, we observed that some candidates relied too heavily on random search; we therefore explicitly discouraged mutations that reduced the construction method to pure stochastic search. In order to score a program, we evaluate the matrices that it produces. Following \citeauthor{nagda2026ramsey} \cite{nagda2026ramsey}, each iteration of program generates two matrices. The first matrix gets a reward proportionate to its number of ones, scaled by a factor of 2 if the amount of ones matches the current $n_{\text{SoTA}}$ and by a factor of 4 if it exceeds it. However, if there are any invalidities in the matrix (a set of any 3 rows and 3 columns such that the intersection contains exclusively ones), the reward is set to -1. The second matrix is treated as a prospect: its reward is inversely proportional to the number of violations in the matrix compared to the number of expected violations given the number of ones.

\begin{algorithm}[t]
\caption{One Phase of Zarankiewicz Program Search}
\label{fig:Alg}
\renewcommand{\algorithmicensure}{\textbf{Output:}}
\begin{algorithmic}[1]
\Require Zarankiewicz parameters $m, n, s, t$
\Ensure Program $p^*$ outputting a graph with the maximum number of 1s

\State \textbf{Init:} Population $\mathcal{P} \leftarrow \{\texttt{p}_{\text{base}}\}$ \Comment{\texttt{p\textsubscript{base}} returns matrices with sparsely populated ones}
\State \textbf{Init:} $n_{\text{SoTA}} \leftarrow \ 0$

\For{\textit{100 iterations}}
    \State $\texttt{p}_{\text{new}} \leftarrow \texttt{LLM\_Mutation}(\texttt{Select}(\mathcal{P}))$
    \State $(M_1, M_2) \leftarrow \texttt{p}_{\text{new}}.\texttt{run}()$
    \Comment{generate 2 new matrices}
    \If{$M_1$ has no violations}
        \State $c_1, c_2 \leftarrow \texttt{count\_ones($M_1$), count\_ones($M_2$)}$
        \State $S_1 \leftarrow \begin{cases} 4 \cdot c_1 & \text{if } c_1 > n_{\text{SoTA}} \\ 2 \cdot c_1 & \text{if } c_1 = n_{\text{SoTA}} \\ c_1 & \text{otherwise} \end{cases}$
        \Else
        \State $S_1 \leftarrow -1$
    \EndIf
    \State $e_{\text{expected}} \leftarrow \dbinom{M}{S}\dbinom{N}{T} \left(\dfrac{c_2}{MN}\right)^{ST}$   
    \State $S_2 \leftarrow \dfrac{1}{2} \max\!\left(0,\, 1 - \dfrac{\texttt{count\_viol}(M_2)}{e_{\text{expected}}}\right)$
    \Statex  where \texttt{count\_viol} returns the number of all-1 submatrices of size $s$,$t$ 

    \State $\texttt{score}(\texttt{p}_{\text{new}}) \leftarrow S_1+S_2$
    \State $\mathcal{P} \leftarrow \mathcal{P} \cup \{(\texttt{p}_{\text{new}}, \texttt{score})\}$
\EndFor
\State \Return $\arg\max_{p \in \mathcal{P}}\, \textit{score}(p)$
\end{algorithmic}
\end{algorithm}

\subsection{Search schedule}
Following empirical experimentation, we tailor a new search schedule detailed below.
For each of our 51 experiments, we run through three phases with 100 iterations each, storing a new checkpoint and best program discovered every 10 iterations. Within each phase, to balance performance and cost, the LLM used for generation is selected at random according to a fixed distribution. In phase one we use Gemini 2.0 Flash \cite{google2025gemini20flash} with 60\% probability and Claude Sonnet 3.7 \cite{anthropic2025claude37sonnet} with 40\% probability. Phase two uses 60\% Gemini 2.0 Flash and 40\% Claude Opus 4.6 \cite{anthropic2026claudeopus46}. Phase three uses 40\% Gemini 2.0 Flash and 60\% Claude Opus 4.6. We found that cheaper models (Flash, Sonnet) were effective in designing a base program, but creating optimizations that brought the lower bound closer to the tight upper bound required more state of the art models.

After phase three, we determine whether further progress was made by comparing the final value of $n_{\text{SoTA}}$ to its value at the end of phase two. If $n_{\text{SoTA}}$ improved and the upper bound had not yet been matched, we run an additional phase of 100 iterations for the corresponding $(m,n)$ tuple, using the same LLM distribution as in phase three: 40\% Gemini 2.0 Flash and 60\% Claude Opus 4.6. We repeat this process until either the upper bound is matched or an entire phase is completed with no further improvement.

\section{Results}
\begin{figure}[h]
    \centering
    \begin{subfigure}{0.32\textwidth}
        \includegraphics[width=\linewidth]{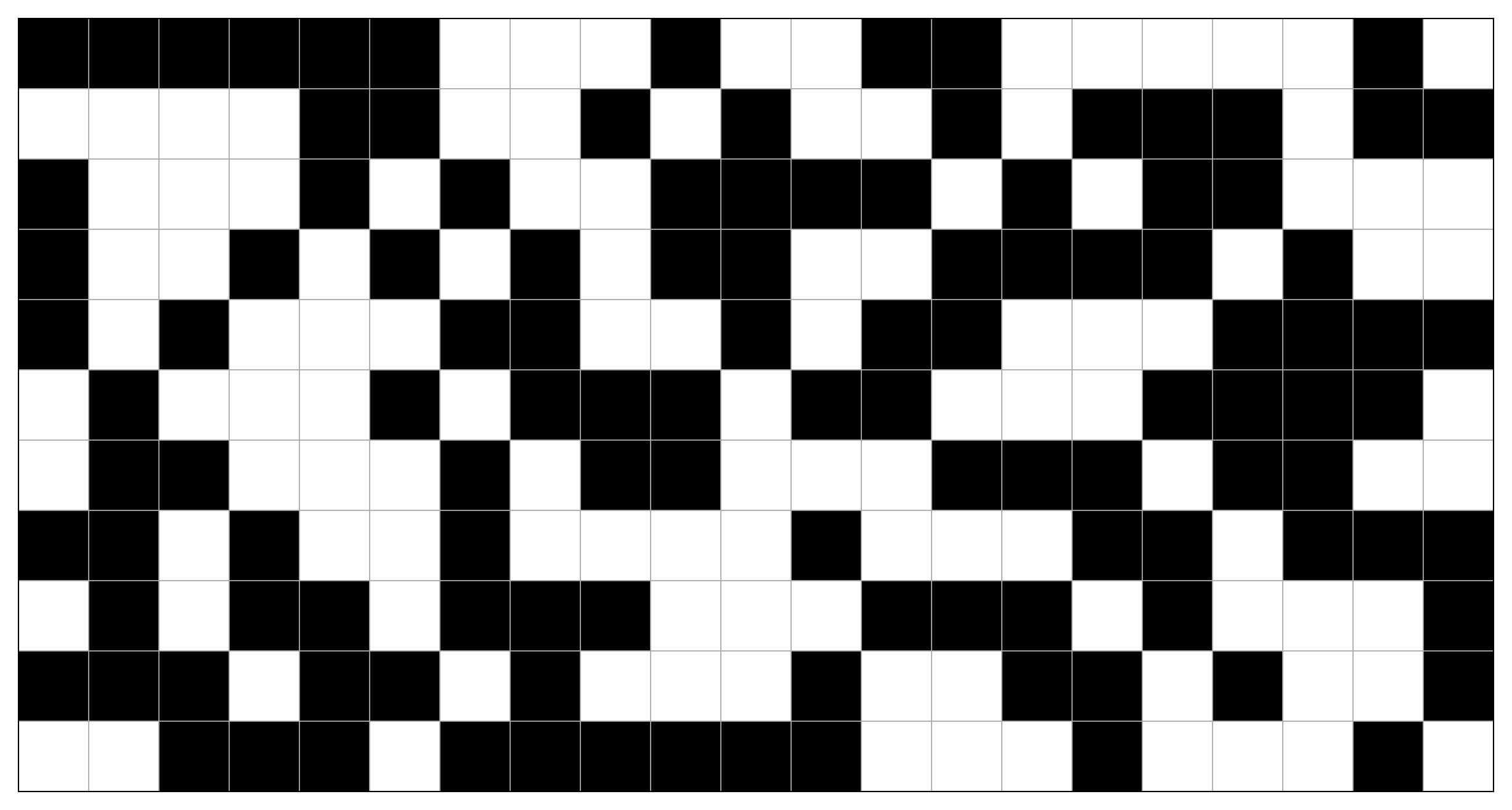}
        \caption{$\textbf{Z}(11, 21, 3, 3)=116$}
    \end{subfigure}
    \hfill
    \begin{subfigure}{0.32\textwidth}
        \includegraphics[width=\linewidth]{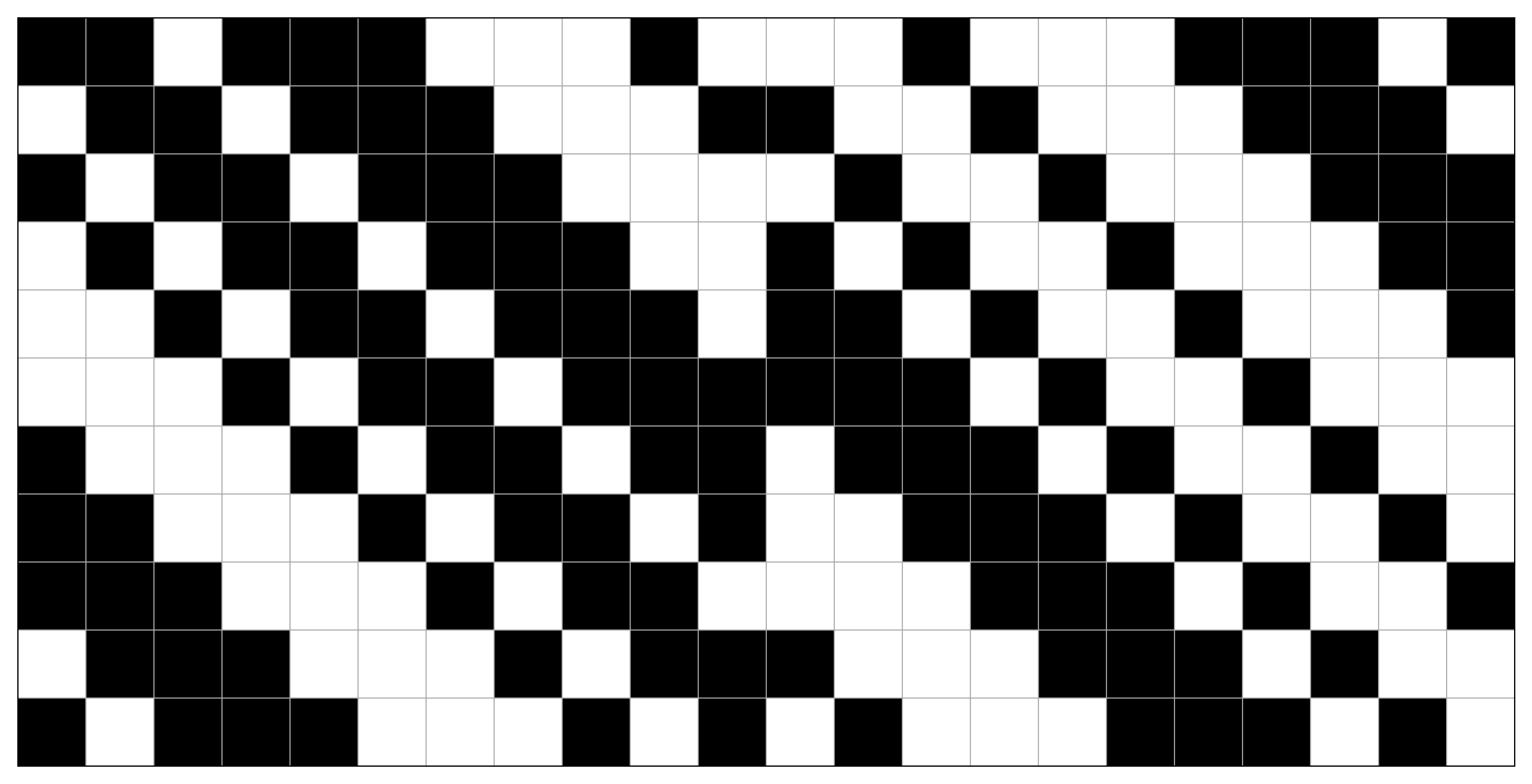}
        \caption{$\textbf{Z}(11, 22, 3, 3)=121$}
    \end{subfigure}
    \hfill
    \begin{subfigure}{0.32\textwidth}
        \includegraphics[width=\linewidth]{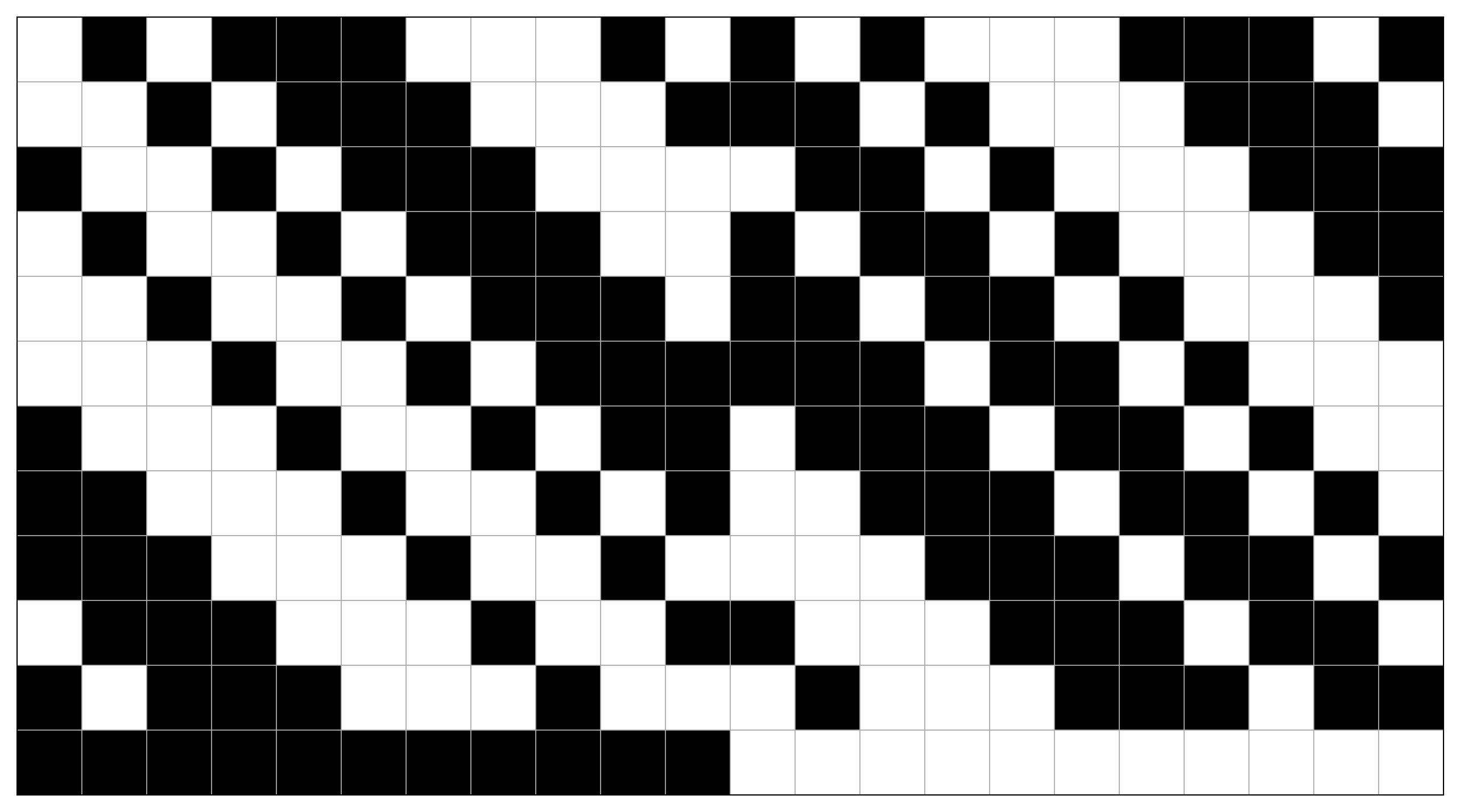}
        \caption{$\textbf{Z}(12, 22, 3, 3)=132$}
    \end{subfigure}
    \caption{Optimal Zarankiewicz matrices for $\textbf{Z}(11, 21, 3, 3)=116$, $\textbf{Z}(11, 22, 3, 3)=121$ and $\textbf{Z}(12, 22, 3, 3)=132$ found by our work. Black squares represent ones and white squares represent zeros.}
    \label{fig:optimal_matrices}
\end{figure}

\begin{figure}
  \centering
  \includegraphics[width=15cm]{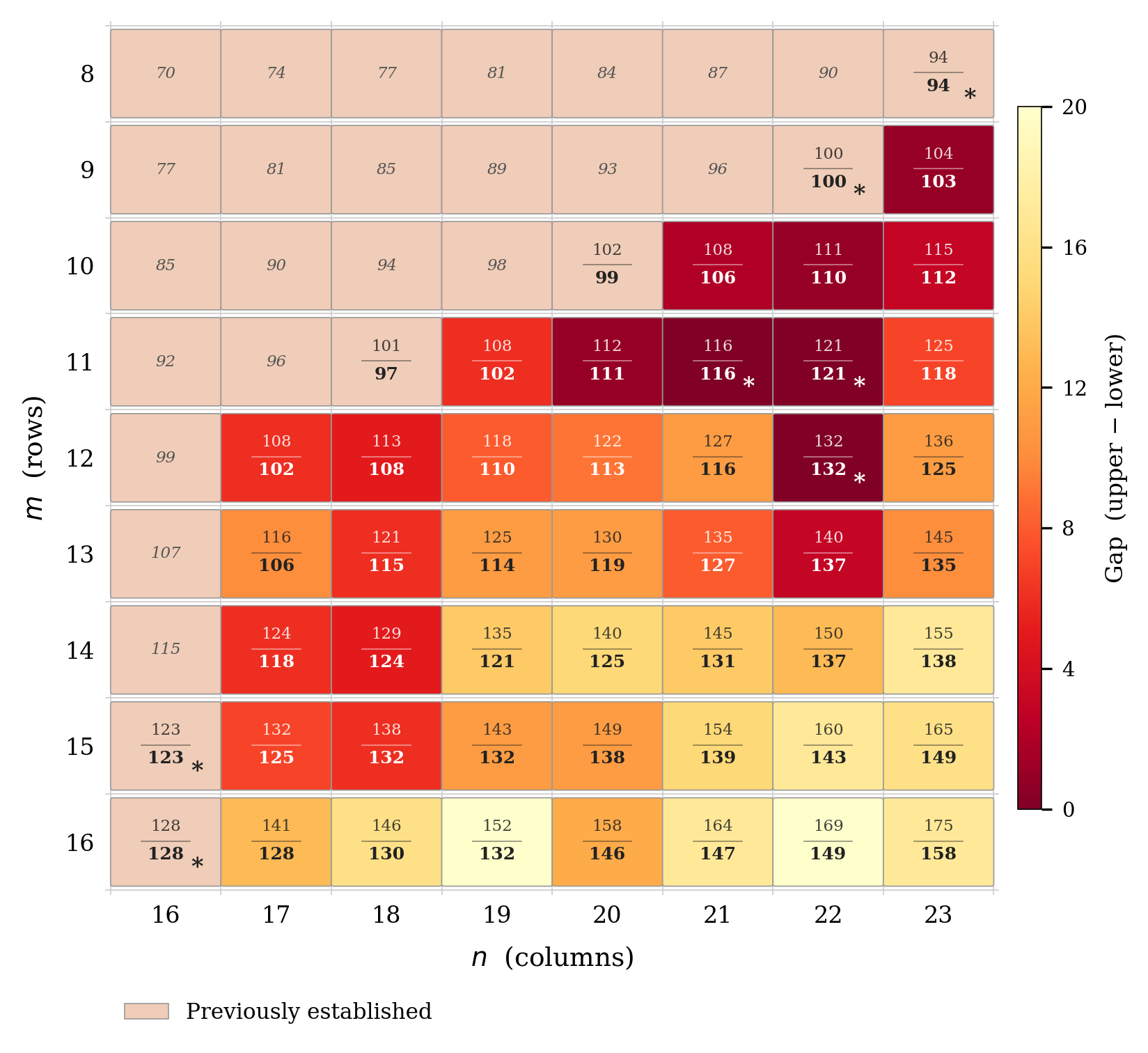} 
  \caption{Upper and lower bounds for Zarankiewicz numbers $\textbf{Z}(m, n, 3, 3)$ with $8 \leq m \leq 16$ and $16 \leq n \leq 23$. The top number shows the previously known upper bound and the bottom number shows the lower bound established by our methods. Asterisks indicate tight cases, where our lower bound matches the best known upper bound. We did not run experiments on established cases with only one number recorded.}
  \label{fig:zar_nums}
\end{figure}

We establish lower bounds for the Zarankiewicz number with $s=t=3$ for 51 different combinations of $m$ and $n$. Out of these 51 different combinations, seven of them already had the exact Zarankiewicz number confirmed, while the other 44 had only an upper bound known. Our method replicates lower bound constructions for four out of the seven previously determined Zarankiewicz numbers and provides constructions proving the upper bounds of three Zarankiewicz numbers tight: $\textbf{Z}(11, 21, 3, 3)=116$, $\textbf{Z}(11, 22, 3, 3)=121$, and $\textbf{Z}(12, 22, 3, 3)=132$. We establish lower bounds for the other 41 out of 44 open cases, some within one away from the upper bound. We report a visualization of the optimal constructions for $\textbf{Z}(11, 21, 3, 3)$, $\textbf{Z}(11, 22, 3, 3)$, and $\textbf{Z}(12, 22, 3, 3)$ in \autoref{fig:optimal_matrices}. A table summarizing the new bounds we determined can be found in \autoref{fig:zar_nums}. It is important to note that many of the lower bounds we found may in fact be optimal, with any remaining gap arising from the known upper bounds not being tight rather than from suboptimality of our constructions.

\section{Cost analysis}
Although many works using AI for scientific discovery report algorithmic performance, the associated costs are often omitted, making it difficult to assess their accessibility. Our work is remarkably inexpensive: We estimate that each $m \times n$ case costed approximately \$15 to \$30 depending on how many iterations the model needed to go through prior to convergence. Moreover, the experiments are also extremely time efficient, with every phase taking around 10 minutes. 
We thus want to remark that LLM based evolutionary discovery methods do not require large scale computational budgets and can instead serve as accessible tools for generating mathematical constructions that lead to new results.

\section{Prior attempts on Zarankiewicz Numbers} 
Determining Zarankiewicz numbers has been an active line of research since Zarankiewicz’s original problem formulated in 1951 \cite{Zarankiewicz1951}. \citet{roman1975zarankiewicz} proved that
\begin{equation}
    \textbf{Z}(m, n, s, t) \leq \frac{(t-1)\binom{m}{s}}{\binom{k}{s-1}} + \frac{(k+1)(s-1)}{s} n
\end{equation}
using a linear program. Later, \citet{tan2022zarankiewiczsat} developed a computational solver by representing the Zarankiewicz number search as a satisfiability (SAT) problem to establish lower bounds, and as an unsatisfiability (UNSAT) problem to establish upper bounds for graphs up to a certain size. Just a short time after, \citet{davies2026zarankiewicz} improved the constraints on the linear program of \citet{roman1975zarankiewicz}, creating an upper bound that improved upon the upper bounds found by \citet{roman1975zarankiewicz} in 29 out of the 44 open cases we focus on \cite{davies2026zarankiewicz}. We cite \citet{tan2022zarankiewiczsat} and \citet{davies2026zarankiewicz} in particular because they are the sources for the values used in this paper, including the upper bounds for $s, t = 3$ against which we verify that our lower bounds are tight.

While most of the literature has focused on developing classical asymptotic upper bounds and on computing exact upper bound values for special cases, the study of lower bounds is still underdeveloped
\cite{conlonZarankiewiczRemarks}. This is partly because lower-bound constructions are often difficult to obtain analytically and frequently require computational or highly problem-specific methods \cite{tan2022zarankiewiczsat}. To our knowledge, this work is the first to provide explicit lower-bound values for the parameter combinations considered.

\section{Generated Search Algorithms}
\label{sec6}
We present the algorithms that yielded constructions proving the tight upper bounds for $\mathbf{Z}(11,21,3,3)$, $\mathbf{Z}(11,22,3,3)$, and $\mathbf{Z}(12,22,3,3)$ in \autoref{alg:11_21}, \autoref{alg:11_22}, and \autoref{alg:12_22}, respectively. We also include the algorithms that reproduced known lower-bound constructions for $\mathbf{Z}(8,23,3,3)$, $\mathbf{Z}(9,22,3,3)$, $\mathbf{Z}(15,16,3,3)$, and $\mathbf{Z}(16,16,3,3)$ in \autoref{alg:8_23}, \autoref{alg:9_22}, \autoref{alg:16_15}, and \autoref{alg:16_16}. All algorithms are reported exactly as generated by our method.

The generated optimal algorithms leading to  fall into three broad categories. Two of the seven algorithms perform minimal computation beyond directly outputting an explicit matrix. 
Two others exploit circulant structure: one uses a circulant block construction algorithm obtained by enumerating modular shift sets, while the other combines a circulant base construction with a greedy construction step. The remaining three algorithms perform randomized perturbation and repair. They: 
\begin{enumerate}
    \item Iteratively improve upon an explicit $m \times n$ starting matrix.
    \item Remove a random number of ones from the matrix before greedily adding them back using a predetermined order.
    \item Create verification functions, both locally (checking whether adding a new one in a certain location would introduce a violation) and globally (ensuring the validity of the matrix in its current form). 
\end{enumerate}

In particular, we note that the latter ripup and repair strategy suggests a potentially more general search framework for constructing dense $K_{s,t}$-free matrices, and may extend to other values of $\mathbf{Z}(m,n,s,t)$.

\section{Discussion and Conclusion}
In this work, we show that evolutionary program search guided by LLMs can produce new extremal graph constructions to prove lower bounds on the Zarankiewicz problem. Our method proves the exact values $\mathbf{Z}(11,21,3,3)=116$, $\mathbf{Z}(11,22,3,3)=121$, and $\mathbf{Z}(12,22,3,3)=132$, establishes lower bounds for 41 further parameter combinations, and reproduces known optimal constructions in four other cases. 

Our results demonstrate that evolutionary program search can be a practical tool for discovering dense $K_{s,t}$-free bipartite graphs in finite regimes where analytic constructions are difficult to obtain.
Further, our experiments are efficient and accessible with respect to both cost and time, with each parameter combination costing less than \$30, and the search completing in the order of minutes. Thus, in problems where progress depends on finding explicit finite constructions, this method may serve as a useful complement to current approaches.

There is more to study about the application of automatic discovery to the Zarankiewicz numbers. We list three ideas for future work below.
\begin{enumerate}
    \item The method can be extended to other combinations of $m$ and $n$ and beyond the diagonal case $s=t=3$ to determine lower bounds and potentially prove exact Zarankiewicz numbers for other parameter combinations. 
    \item The ripup and repair strategies discussed in \ref{sec6} could be analyzed and consolidated into a universal lower-bound search procedure for Zarankiewicz numbers.
    \item Finally, we believe that discovery via evolutionary algorithms can be extended beyond generating constructions to automatically proving upper bounds.

\end{enumerate}

\bibliographystyle{plainnat}
\bibliography{references}

\appendix
\section{Successful Search Algorithms }

\captionof{algorithm}{Search algorithm establishing for the first time that the previously known upper bound for $\mathbf{Z}(11,21,3,3)$ is tight.}
\label{alg:11_21}
\begin{lstlisting}[style=pythonstyle]
import numpy as np

def construct_graphs():
    M, N = 11, 21

    def is_valid(G):
        for i in range(M):
            for j in range(i+1, M):
                for k in range(j+1, M):
                    if np.sum(G[i] & G[j] & G[k]) >= 3:
                        return False
        return True

    def can_set(G, r, c):
        G[r, c] = 1
        for i in range(M):
            if i == r or G[i, c] == 0:
                continue
            for j in range(i+1, M):
                if j == r or G[j, c] == 0:
                    continue
                if np.sum(G[r] & G[i] & G[j]) >= 3:
                    G[r, c] = 0
                    return False
        return True

    # Seed from the best known valid matrix (111 ones)
    base = np.array([
        [1,1,1,1,1,1,0,0,0,1,0,0,1,0,0,0,0,0,1,1,0],
        [1,0,0,0,1,1,0,0,1,0,1,1,0,1,0,1,0,0,0,1,1],
        [1,0,0,0,1,0,1,0,0,1,1,1,1,0,1,0,1,1,0,0,0],
        [1,0,0,1,0,1,0,1,0,1,0,0,0,1,1,1,1,0,1,0,0],
        [1,0,1,0,0,0,1,1,0,0,1,0,1,1,0,0,0,1,1,1,1],
        [0,1,0,0,0,1,0,1,1,1,0,1,1,0,0,0,1,1,0,1,0],
        [0,1,1,0,0,0,1,0,1,1,1,1,0,1,1,1,0,0,1,0,0],
        [0,1,0,1,0,0,1,1,0,0,0,1,0,0,0,1,1,0,1,1,1],
        [0,1,0,1,1,0,1,0,1,0,0,0,1,1,1,0,1,0,0,0,1],
        [0,1,1,0,1,1,0,1,0,0,1,0,0,0,1,1,0,1,0,0,1],
        [0,0,1,1,1,0,1,0,1,1,0,0,0,0,0,1,0,1,0,1,0],
    ], dtype=int)

    best_G = base.copy()
    best_count = int(np.sum(base))

    for seed in range(300):
        rng = np.random.RandomState(seed * 53 + 17)
        G = base.copy()

        # Try removing k ones and refilling
        ones = list(zip(*np.where(G == 1)))
        rng.shuffle(ones)
        k = rng.randint(3, 12)
        for idx in range(min(k, len(ones))):
            G[ones[idx][0], ones[idx][1]] = 0

        order = [(i, j) for i in range(M) for j in range(N)]
        rng.shuffle(order)
        for i, j in order:
            if G[i, j] == 0:
                can_set(G, i, j)

        for iteration in range(200):
            improved = False
            ones_list = list(zip(*np.where(G == 1)))
            rng.shuffle(ones_list)
            for oi, oj in ones_list:
                G[oi, oj] = 0
                added = []
                zeros = list(zip(*np.where(G == 0)))
                rng.shuffle(zeros)
                for ni, nj in zeros:
                    if can_set(G, ni, nj):
                        added.append((ni, nj))
                        if len(added) >= 2:
                            break
                if len(added) >= 2:
                    improved = True
                else:
                    for ai, aj in added:
                        G[ai, aj] = 0
                    G[oi, oj] = 1
            for i in range(M):
                for j in range(N):
                    if G[i, j] == 0:
                        can_set(G, i, j)
            if not improved:
                break

        count = int(np.sum(G))
        if count > best_count and is_valid(G):
            best_count = count
            best_G = G.copy()
            if best_count >= 116:
                break

    G1 = best_G
    rng2 = np.random.RandomState(999)
    G2 = G1[:, rng2.permutation(N)][rng2.permutation(M), :].copy()
    return G1, G2

def run_graph():
    return construct_graphs()
\end{lstlisting}

\captionof{algorithm}{Search algorithm establishing for the first time that the previously known upper bound for $\mathbf{Z}(11,22,3,3)$ is tight.}
\label{alg:11_22}
\begin{lstlisting}[style=pythonstyle]
M = 11
N = 22
S = 3
T = 3

# EVOLVE-BLOCK-START
import numpy as np
from itertools import combinations

def construct_graphs():
    # Circulant construction: two blocks, each 11x11 circulant
    # For K_{3,3}-free: any 3 rows share at most 2 common 1-columns
    # Use sets where triple intersection of shifted copies is <= 2

    # Block 1: shift set S1, Block 2: shift set S2
    # Row i, col j in block b: entry = 1 iff (j-i)%11 in S_b

    # Best found: S1 with 7 elements, S2 with 7 elements = 154 ones
    # Need: for any 3 distinct rows (a,b,c),
    def check_valid(S1, S2):
        for d1 in range(1, 11):
            for d2 in range(d1+1, 11):
                s1_set = set(S1)
                c1 = len(
                    s1_set
                    & {(x+d1)%11 for x in s1_set}
                    & {(x+d2)%11 for x in s1_set}
                )
                s2_set = set(S2)
                c2 = len(
                    s2_set
                    & {(x+d1)%11 for x in s2_set}
                    & {(x+d2)%11 for x in s2_set}
                )
                if c1 + c2 >= 3:
                    return False
        return True

    best = ([], [], 0)
    for s1_size in range(8, 4, -1):
        for S1 in combinations(range(11), s1_size):
            for s2_size in range(min(22-s1_size*11//11, 8), 4, -1):
                for S2 in combinations(range(11), s2_size):
                    if (len(S1)+len(S2))*11 <= best[2]:
                        continue
                    if check_valid(S1, S2):
                        total = (len(S1)+len(S2))*11
                        if total > best[2]:
                            best = (S1, S2, total)
                            if total >= 154:
                                break
                if best[2] >= 154:
                    break
            if best[2] >= 154:
                break
        if best[2] >= 154:
            break

    S1, S2 = best[0], best[1]
    G1 = np.zeros((M, N), dtype=int)
    for i in range(11):
        for j in range(11):
            if (j - i) % 11 in S1:
                G1[i, j] = 1
            if (j - i) % 11 in S2:
                G1[i, j + 11] = 1

    G2 = G1.copy()
    return G1, G2

def run_graph():
    return construct_graphs()
# EVOLVE-BLOCK-END

if __name__ == "__main__":
    G1, G2 = run_graph()
    print(f"G1 ones: {G1.sum()}, row sums: {G1.sum(axis=1).tolist()}")
\end{lstlisting}

\captionof{algorithm}{Search algorithm establishing for the first time that the previously known upper bound for $\mathbf{Z}(12,22,3,3)$ is tight.}
\label{alg:12_22}
\begin{lstlisting}[style=pythonstyle]
M = 12  # number of rows
N = 22  # number of columns
S = 3   # no K_{S,T} subgraph allowed
T = 3

# EVOLVE-BLOCK-START
import numpy as np

def construct_graphs():
    # Circulant construction on GF(11) + 1 extra row
    # Block1 (cols 0-10): row i has 1s at (i+d)%11 for d in D1
    # Block2 (cols 11-21): row i has 1s at (i+d)%11 for d in D2
    # D1 and D2 chosen so any 2 rows share <= 2 cols in each block
    # and cross-block overlaps stay controlled
    D1 = [1, 3, 4, 5, 9]  # QR mod 11, |D1| = 5
    D2 = [0, 2, 6, 7, 8, 10]  # complement, |D2| = 6

    G = np.zeros((M, N), dtype=int)
    for i in range(11):
        for d in D1:
            G[i, (i + d) % 11] = 1
        for d in D2:
            G[i, 11 + (i + d) % 11] = 1

    # Row 11: greedy fill
    from itertools import combinations

    def valid(mat):
        for r in combinations(range(M), 3):
            if np.sum(mat[r[0]] & mat[r[1]] & mat[r[2]]) >= 3:
                return False
        return True

    for j in range(N):
        G[11, j] = 1
        if not valid(G):
            G[11, j] = 0

    # Greedy improve all cells
    for i in range(M):
        for j in range(N):
            if G[i, j] == 0:
                G[i, j] = 1
                if not valid(G):
                    G[i, j] = 0

    # Second pass in reverse order for potentially different improvements
    for i in range(M - 1, -1, -1):
        for j in range(N - 1, -1, -1):
            if G[i, j] == 0:
                G[i, j] = 1
                if not valid(G):
                    G[i, j] = 0

    G2 = G.copy()
    return G, G2

def run_graph():
    np.random.seed(42)
    return construct_graphs()
# EVOLVE-BLOCK-END

if __name__ == "__main__":
    G1, G2 = run_graph()
    print(f"G1 shape: {G1.shape}, ones: {G1.sum()}, ones/row: {G1.sum(axis=1).tolist()}")
    print(f"G2 shape: {G2.shape}, ones: {G2.sum()}")
    print("G1:\n", G1)
\end{lstlisting}

\captionof{algorithm}{Search algorithm obtaining the construction that reproduces the known exact value for $\mathbf{Z}(8,23,3,3)$.}
\label{alg:8_23}
\begin{lstlisting}[style=pythonstyle]
M = 8
N = 23
S = 3
T = 3

# EVOLVE-BLOCK-START
import numpy as np

def run_graph():
    # Verified 94-one K_{3,3}-free matrix
    # Row degrees: [11,12,12,12,11,12,12,12] = 94
    # Every triple of rows shares at most 2 common columns
    G1 = np.array([
        [0,0,0,0,1,1,0,1,1,0,1,0,1,1,1,0,0,1,1,0,1,0,0],
        [1,0,1,1,0,1,1,1,1,0,1,0,0,1,0,1,0,0,0,1,0,0,1],
        [0,1,0,1,1,1,1,0,0,0,0,0,0,1,0,0,1,0,1,1,1,1,1],
        [1,0,1,1,1,1,0,1,0,0,0,1,1,0,1,1,1,0,0,0,0,1,0],
        [1,1,0,0,1,0,0,0,1,1,0,0,0,1,1,1,0,1,0,1,0,1,0],
        [0,0,1,0,0,0,0,1,1,1,0,1,0,0,0,0,1,1,1,1,1,1,1],
        [0,1,1,0,1,0,1,0,0,1,1,1,1,0,0,1,0,1,0,0,1,0,1],
        [1,1,0,1,0,0,1,0,1,1,1,1,1,0,1,0,1,0,1,0,0,0,0],
    ], dtype=int)
    # Set G2 = G1 to try for symmetry/matching bonus
    G2 = G1.copy()
    return G1, G2
# EVOLVE-BLOCK-END

if __name__ == "__main__":
    from itertools import combinations
    G1, G2 = run_graph()
    total = int(G1.sum())
    print(f"G1 ones: {total}, row sums: {G1.sum(axis=1).tolist()}")
    valid = True
    for r1, r2, r3 in combinations(range(M), 3):
        common = sum(1 for c in range(N) if G1[r1,c] and G1[r2,c] and G1[r3,c])
        if common >= 3:
            print(f"VIOLATION: rows {r1},{r2},{r3} share {common} columns")
            valid = False
    print(f"Valid: {valid}")
\end{lstlisting}

\captionof{algorithm}{Search algorithm obtaining the construction that reproduces the known exact value for $\mathbf{Z}(9,22,3,3)$.}
\label{alg:9_22}
\begin{lstlisting}[style=pythonstyle]
M = 9  # number of rows
N = 22  # number of columns
S = 3   # no K_{S,T} subgraph allowed
T = 3

# EVOLVE-BLOCK-START
import numpy as np
from itertools import combinations

def construct_graphs():
    def is_valid(mat):
        for r1, r2, r3 in combinations(range(mat.shape[0]), 3):
            if np.sum(mat[r1] & mat[r2] & mat[r3]) >= 3:
                return False
        return True

    def triple_violations(mat):
        count = 0
        for r1, r2, r3 in combinations(range(mat.shape[0]), 3):
            c = np.sum(mat[r1] & mat[r2] & mat[r3])
            if c >= 3:
                count += c - 2
        return count

    def can_add(mat, i, j):
        others = [r for r in range(M) if r != i and mat[r, j] == 1]
        for r1, r2 in combinations(others, 2):
            common = 0
            for k in range(N):
                if mat[i, k] and mat[r1, k] and mat[r2, k]:
                    common += 1
                    if common >= 3:
                        return False
        return True

    def greedy_fill(mat):
        mat = mat.copy()
        changed = True
        while changed:
            changed = False
            cells = [(mat[i].sum(), i, j) for i in range(M) for j in range(N) if mat[i,j]==0]
            cells.sort()
            for _, i, j in cells:
                if mat[i,j] == 0:
                    mat[i,j] = 1
                    if can_add(mat, i, j):
                        changed = True
                    else:
                        mat[i,j] = 0
        return mat

    # Exhaustive SA-based search from multiple seeds
    best_mat = None
    best_count = 0

    rows97 = [
        [1,1,0,0,0,0,1,1,0,0,0,0,0,0,1,0,0,1,1,1,1,1],
        [1,1,1,1,0,1,0,0,1,1,0,1,0,1,0,0,0,1,0,0,1,1],
        [0,1,0,1,1,1,0,0,0,1,1,0,1,0,0,0,0,0,1,1,0,1],
        [0,0,0,1,0,0,1,1,0,0,0,1,1,0,0,1,1,1,1,0,0,1],
        [1,0,0,0,1,1,0,0,1,0,0,0,0,1,1,1,1,0,1,0,0,1],
        [0,0,1,0,0,1,1,0,0,1,1,0,1,1,0,1,1,1,0,1,1,0],
        [0,0,1,0,1,0,0,1,1,0,1,1,1,0,1,0,1,0,0,0,1,1],
        [1,1,1,1,1,0,1,1,0,0,1,1,0,1,1,1,0,0,0,1,0,0],
        [0,1,0,0,1,0,0,1,1,1,0,0,1,0,0,1,0,0,1,0,1,0],
    ]
    base = np.array(rows97, dtype=int)

    np.random.seed(31415)
    for seed_trial in range(20):
        current = base.copy()
        perm = np.random.permutation(N)
        current = current[:, perm]
        current = greedy_fill(current)
        cur_count = int(current.sum())

        for trial in range(2000):
            mat = current.copy()
            ones = list(zip(*np.where(mat==1)))
            np.random.shuffle(ones)
            n_rem = np.random.randint(2, 10)
            for k in range(min(n_rem, len(ones))):
                mat[ones[k]] = 0
            mat = greedy_fill(mat)
            c = int(mat.sum())
            if c >= cur_count:
                current = mat
                cur_count = c
            if c > best_count:
                best_count = c
                best_mat = mat.copy()
                if c >= 100:
                    break
        if best_count >= 100:
            break

    G1 = best_mat if best_mat is not None else greedy_fill(base)
    G2 = np.ones((M, N), dtype=int)
    return G1, G2

def run_graph():
    return construct_graphs()
# EVOLVE-BLOCK-END

if __name__ == "__main__":
    G1, G2 = run_graph()
    print(f"G1 shape: {G1.shape}, ones: {G1.sum()}, ones/row: {G1.sum(axis=1).tolist()}")
\end{lstlisting}

\captionof{algorithm}{Search algorithm obtaining the construction that reproduces the known exact value for $\mathbf{Z}(15,16,3,3)$.}
\label{alg:16_15}
\begin{lstlisting}[style=pythonstyle]
M = 16
N = 15
S = 3
T = 3

# EVOLVE-BLOCK-START
import numpy as np

def construct_graphs():
    # Exact best known 16x15 matrix with 123 ones, no 3x3 all-ones submatrix
    rows = [
        [1,1,1,1,1,1,1,1,0,0,0,0,0,0,0],  # r0: deg 8
        [1,1,1,1,0,0,0,0,1,1,1,1,0,0,0],  # r1: deg 8
        [1,1,0,0,1,1,0,0,1,1,0,0,1,1,0],  # r2: deg 8
        [1,1,0,0,0,0,1,1,0,0,1,1,1,1,0],  # r3: deg 8
        [0,0,1,1,1,1,0,0,0,0,1,1,1,1,0],  # r4: deg 8
        [0,0,1,1,0,0,1,1,1,1,0,0,1,1,0],  # r5: deg 8
        [0,0,0,0,1,1,1,1,1,1,1,1,0,0,0],  # r6: deg 8
        [1,0,1,0,1,0,1,0,1,0,1,0,1,0,1],  # r7: deg 8
        [1,0,1,0,0,1,0,1,0,1,0,1,1,0,1],  # r8: deg 8
        [1,0,0,1,1,0,0,1,0,1,1,0,0,1,1],  # r9: deg 8
        [1,0,0,1,0,1,1,0,1,0,0,1,0,1,1],  # r10: deg 8
        [0,1,1,0,1,0,0,1,1,0,0,1,0,1,1],  # r11: deg 8
        [0,1,1,0,0,1,1,0,0,1,1,0,0,1,1],  # r12: deg 8
        [0,1,0,1,1,0,1,0,0,1,0,1,1,0,1],  # r13: deg 8
        [0,1,0,1,0,1,0,1,1,0,1,0,1,0,1],  # r14: deg 8
        [0,0,0,0,1,1,0,0,0,0,0,0,0,0,1],  # r15: deg 3
    ]
    G1 = np.array(rows, dtype=int)

    # Build G2: add one extra 1 per row where possible
    G2 = G1.copy()
    from itertools import combinations

    def is_valid(mat):
        for r1, r2, r3 in combinations(range(M), 3):
            common = mat[r1] & mat[r2] & mat[r3]
            if np.where(common)[0].shape[0] >= 3:
                return False
        return True

    for i in range(M):
        zeros = np.where(G2[i] == 0)[0]
        for j in zeros:
            G2[i, j] = 1
            if not is_valid(G2):
                G2[i, j] = 0
            else:
                break

    return G1, G2

def run_graph():
    return construct_graphs()
# EVOLVE-BLOCK-END

if __name__ == "__main__":
    G1, G2 = run_graph()
    print(f"G1 ones: {G1.sum()}, per row: {G1.sum(axis=1).tolist()}")
    print(f"G2 ones: {G2.sum()}")
    print("G1:\n", G1)
\end{lstlisting}

\captionof{algorithm}{Search algorithm obtaining the construction that reproduces the known exact value for $\mathbf{Z}(16,16,3,3)$.}
\label{alg:16_16}
\begin{lstlisting}[style=pythonstyle]
M = 16
N = 16
S = 3
T = 3

# EVOLVE-BLOCK-START
import numpy as np

def construct_graphs():
    def check_add(G, i, j):
        ri = [r for r in range(16) if r != i and G[r, j]]
        ci = [c for c in range(16) if c != j and G[i, c]]
        if len(ri) < 2 or len(ci) < 2:
            return True
        for a in range(len(ri)):
            for b in range(a+1, len(ri)):
                if sum(1 for c in ci if G[ri[a], c] and G[ri[b], c]) >= 2:
                    return False
        for a in range(len(ci)):
            for b in range(a+1, len(ci)):
                if sum(1 for r in ri if G[r, ci[a]] and G[r, ci[b]]) >= 2:
                    return False
        return True

    def is_valid(G):
        for r1 in range(16):
            for r2 in range(r1+1, 16):
                common = np.where(G[r1] & G[r2])[0]
                if len(common) < 3:
                    continue
                for r3 in range(r2+1, 16):
                    if np.sum(G[r3, common]) >= 3:
                        return False
        return True

    # Best known 119-ones matrix from previous run - use as seed
    R = [
        [1,1,0,0,1,0,0,0,0,0,1,1,0,1,0,1],
        [1,1,1,0,0,1,0,0,0,0,0,1,1,0,1,1],
        [0,1,1,1,0,0,1,0,0,0,0,0,1,1,0,1],
        [1,0,1,1,1,0,0,1,0,0,0,0,0,1,1,0],
        [0,1,0,1,1,1,0,0,1,0,0,0,0,0,1,1],
        [1,0,1,1,1,1,1,0,0,1,0,0,0,0,0,1],
        [1,1,0,1,0,1,1,1,0,0,1,0,0,0,0,0],
        [0,1,1,0,1,1,1,1,1,0,0,1,0,0,0,0],
        [0,0,1,1,0,1,0,1,1,1,0,0,1,0,0,0],
        [0,0,0,1,1,0,1,0,1,1,1,0,0,1,0,0],
        [0,0,0,0,1,1,0,1,0,1,1,1,0,0,1,0],
        [0,0,0,0,0,1,1,0,1,0,1,1,1,0,0,1],
        [1,0,1,0,0,0,1,1,0,1,1,1,1,1,0,0],
        [0,1,0,1,0,0,0,1,1,0,1,0,1,1,1,0],
        [0,0,1,0,1,0,0,0,1,1,0,1,0,1,1,1],
        [1,0,0,1,0,0,0,0,0,1,1,0,1,0,1,1]
    ]
    best = np.array(R, dtype=int)
    best_sum = best.sum() if is_valid(best) else 0

    for trial in range(100):
        rng = np.random.RandomState(trial + 200)
        G = best.copy()
        ones = list(zip(*np.where(G == 1)))
        rng.shuffle(ones)
        for _ in range(rng.randint(2, 8)):
            if ones:
                idx = rng.randint(len(ones))
                G[ones[idx]] = 0
        zeros = [(i, j) for i in range(16) for j in range(16) if G[i, j] == 0]
        rng.shuffle(zeros)
        for i, j in zeros:
            if G[i, j] == 0 and check_add(G, i, j):
                G[i, j] = 1
        if is_valid(G) and G.sum() > best_sum:
            best_sum = G.sum()
            best = G.copy()

    G2 = best.copy()
    return best, G2

def run_graph():
    return construct_graphs()
# EVOLVE-BLOCK-END

if __name__ == "__main__":
    G1, G2 = run_graph()
    print(f"G1 ones:{G1.sum()}, density:{G1.sum()/256:.4f}")
    print(G1)
\end{lstlisting}

\end{document}